# VQ-VAD: VECTOR-QUANTIZED MOTION REPRESENTATION LEARNING FOR HUMAN-CENTRIC VIDEO ANOMALY DETECTION

Narges Rashvand, Ghazal Alinezhad Noghre, Shanle Yao, Gabriel Maldonado, Hamed Tabkhi
Electrical and Computer Engineering Department
*University of North Carolina at Charlotte*
Charlotte, NC, USA
{nrashvan, galinezh, syao, gmaldon2, htabkhiv}@charlotte.edu

**ABSTRACT**

Video Anomaly Detection (VAD) is inherently challenging due to the scarcity of anomalies and the large visual variability in surveillance footage, including changes in lighting, viewpoint, and human appearance. To mitigate visual noise and address privacy concerns, recent work has shifted to pose-based VAD, which focuses on motion dynamics rather than raw video data. However, existing pose-based approaches model human behavior in continuous latent spaces, limiting their ability to learn compact motion patterns necessary for robust behavior analysis. We address this by proposing Vector-Quantized Video Anomaly Detection (VQ-VAD), a novel human-centric anomaly detection framework that learns discrete motion representations. VQ-VAD adapts Vector-Quantized GAN (VQ-GAN), originally developed for image generation, to operate on keypoint sequences and construct a motion codebook of normal behavior. Trained exclusively on normal motion sequences, VQ-VAD detects anomalies by identifying high reconstruction errors when an observed motion sequence cannot be mapped to the learned codebook. We conduct extensive experiments across three complementary evaluation settings, including in-domain, cross-domain, and cross-dataset generalization, on four anomaly detection benchmarks. VQ-VAD achieves strong in-domain accuracy (81.83% on HR-SHT [15]), effective cross-domain transfer from CMU Panoptic [14] (76.69% on HR-SHT [15] without retraining), and competitive cross-dataset robustness. The code base for this work is available at https://github.com/TeCSAR-UNCC/VQ-VAD.

## 1. INTRODUCTION

Video anomaly detection (VAD) aims to identify observations that deviate from normal visual patterns and is widely used in applications such as public safety, traffic analysis, and healthcare [2, 6, 20, 24, 25]. Human-centric VAD [2, 6, 11, 12, 13, 19, 20] focuses specifically on anomalies in human behavior and motion, emphasizing the semantic understanding of actions, poses, and interactions in applications such as crowd surveillance, patient monitoring, and workplace observation.

Human-centric VAD is commonly approached through two paradigms: pixel-based and pose-based methods. Pixel-based approaches [26] operate directly on raw visual data, leveraging spatiotemporal cues from video frames. While this enables the use of rich contextual information, it also introduces high-dimensional, noisy inputs and can introduce appearance-related biases tied to clothing, lighting, or background [6, 20]. Limited visual diversity also increases false alarms as uncommon but normal appearances may be misclassified as anomalies. These issues have motivated the exploration of more abstract representations. Pose-based methods [11, 12, 13, 19, 20, 26], which rely on skeletal motion rather than pixel intensities, focus on human dynamics to detect behavioral irregularities while preserving privacy and reducing visual bias. Another fundamental challenge in VAD, as in general anomaly detection, lies in the inherent uncertainty of what constitutes an anomaly. Since anomalous behaviors are rare and unpredictable [6], it is often infeasible to collect or define them explicitly. Consequently, recent research has shifted toward self-supervised and semi-supervised paradigms, where models learn normal behavior through proxy tasks and identify deviations as anomalies. While these approaches have advanced the field by reducing the need for labeled anomalies, they remain sensitive to environmental and setup variations such as camera viewpoints, scene geometry, or motion patterns, which can significantly degrade performance [6]. This sensitivity limits their generalizability and leads to increased false alarms when deployed across diverse real-world settings [20]. A key factor in improving generalizability for human-centric VAD lies in how human motion is represented. The human body consists of joints and limbs with constrained degrees of freedom, resulting in a finite set of primitive movements whose combinations form complex actions, much like words forming sentences in language. Drawing inspiration from this analogy, human motion can be modeled within a discrete latent space, enabling the construction of a compact motion vocabulary that encapsulates essential movement semantics. This abstraction reduces noise and visual redundancy while providing a more structured and transferable representation, thereby enhancing robustness across diverse environments.

In this paper, we explore human motion discretization for pose-based video anomaly detection, a direction not previously studied. Inspired by VQ-GANs [9], we propose Vector Quantized Video Anomaly Detection (VQ-VAD), which learns compact discrete motion representations from pose sequences through a learned codebook. This discrete latent space provides a structured and noise-resilient representation for robust anomaly detection. We extensively evaluate VQ-VAD under in-domain, cross-domain, and cross-dataset settings. In cross-domain evaluation, VQ-VAD trained on the CMU Panoptic Dataset [14] achieves 75.09% on SHT [15] without fine-tuning. In cross-dataset evaluation, training on SHT [15] yields stable performance on NWPUC [3] (61.81%) and HuVAD [7] (59.66%), demonstrating strong generalization under domain shift. In summary, the contributions of this work are as follows:

- We introduce a compact and semantically rich representation of human motion by constructing a discrete motion vocabulary, enabling more structured and interpretable motion modeling.
- We propose VQ-VAD, a novel framework that leverages discrete human motion representations to achieve more robust and generalizable video anomaly detection.
- We define two evaluation protocols on top of general in-domain benchmarks, cross-domain generalization, and cross-dataset generalization, to systematically measure robustness.

## 2. RELATED WORKS

Historically, VAD systems were built on handcrafted features [5]. While effective in controlled scenarios, these methods struggled to generalize to real-world environments because of their reliance on predefined feature descriptors. The development of deep learning introduced a major shift in VAD research, leading to two dominant paradigms: pixel-based and pose-based approaches. Pixel-based methods [1, 23] operate directly on raw visual content, achieving strong performance but inheriting limitations such as sensitivity to lighting, scene appearance, and occlusions. Moreover, their reliance on appearance cues raises increasing privacy concerns in human-centric surveillance settings. Pose-based VAD [11, 20, 26], in contrast, abstracts videos into skeletal representations, preserving essential motion dynamics while avoiding identifiable visual information. This abstraction provides robustness against background variations and offers a privacy-preserving alternative.

Semi-supervised and unsupervised pose-based VAD methods commonly learn normal behavioral patterns through self-supervised objectives. These approaches either reconstruct the current pose frame [4, 13, 17] or predict past or future motion sequences [12]. Several studies further combine both objectives within multi-branch architectures [18, 19], demonstrating improved anomaly detection capability through richer motion modeling. Several pose-based anomaly detection methods have been proposed to avoid appearance variability and focus directly on human motion dynamics. Among them, GEPC [17] maps pose sequences into a latent embedding space and detects anomalies via deviations from cluster-consistent representations. STG-NF [11] applies spatio-temporal graph normalizing flows to reshape pose sequences into a standard distribution, where outliers indicate anomalous events. TSGAD [19] incorporates trajectory-aware reasoning through a graph variational autoencoder, using prediction–observation mismatch to derive anomaly scores. While existing pose-based methods primarily rely on continuous latent representations, recent advances in vector-quantized models offer an alternative approach based on discrete representations. VQ-GANs [8] were originally introduced for image-to-image translation and image generation by learning a discrete codebook within a compressed latent space. Drawing inspiration from the success of VQ-GAN in high-fidelity image synthesis, we adopt a VQ-GAN framework to model temporal human pose sequences for robust video anomaly detection.

## 3. METHODOLOGY

VQ-VAD represents human motion using a discrete learned vocabulary of tokens, unlike prior methods that rely on continuous latent spaces. We adopt a VQ-GAN architecture for temporal keypoint sequences to learn a compact latent codebook that captures human motion patterns (Figure 1).

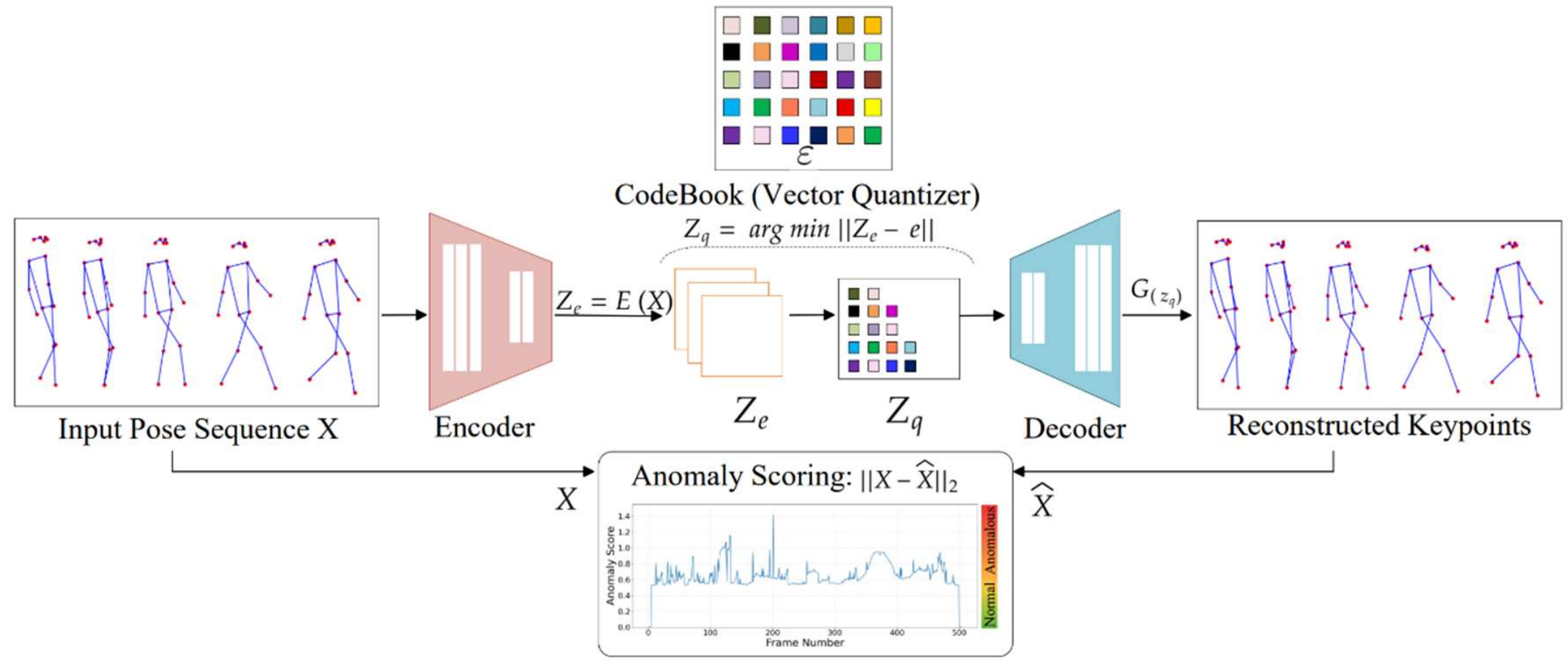

Figure 1. VQ-VAD architecture for unsupervised human anomaly detection based on pose sequences. The input pose sequence X is encoded into a continuous latent representation $Z_e$, quantized via the learned codebook to obtain discrete motion tokens $Z_q$, and reconstructed. Anomaly scores are computed using reconstruction errors.

### 3.1 Pose Sequence Representation and Preprocessing

Each video is represented as a pose sequence $P \in R^{F\times K\times D}$, where $F$ is the number of frames, $K$ is the number of keypoints, and $D \in \{2,3\}$ denotes the spatial dimension. For temporal processing, the keypoint and spatial dimensions are flattened into feature channels, producing $X \in R^{F\times C}$, where $C = K \times D$. This representation enables 1D temporal convolutions to efficiently model motion dynamics.

### 3.2 VQ-VAD Architecture and Training Objective

VQ-VAD uses a vector-quantized autoencoder with adversarial learning to learn a discrete latent space of normal motion dynamics for anomaly detection via reconstruction error.
The VQ-VAD consists of three main components:

- Encoder ($E(.)$): Stacked 1D convolutional blocks encode the input pose sequence $X$ into a continuous latent representation $z_e = E(X)$.
- Vector Quantizer (Codebook): Discretizes the continuous latent space by mapping each latent vector $z_e$ to its nearest neighbor, $e_i$, in the learned codebook which is a finite set of motion embeddings, $\varepsilon = \{e_1, e_2, \ldots, e_k\}$. The quantized latent representation is $Z_q = Quantize\ (z_e) = \arg min_{e_i \in \varepsilon} \|z_e - e_i\|_2$.
- Decoder: The decoder reconstructs the original keypoint sequence from the discrete motion tokens, $\hat{X} = G(z_q)$. During training, a discriminator encourages smooth and realistic temporal motion.

VQ-VAD is trained end-to-end using a combined loss that encourages accurate reconstruction, commitment to discrete codebook entries, and perceptually realistic motion, $\mathcal{L}_{total} = \mathcal{L}_{rec} + \mathcal{L}_{VQ} + \lambda\mathcal{L}_{adv}$, where reconstruction loss ($\mathcal{L}_{rec}$) measures the similarity between the input $X$ and the reconstructed output $\hat{X}$. We utilize Mean Squared Error (MSE) to ensure reconstruction accuracy, defined as $\mathcal{L}_{rec} = \|X - \hat{X}\|_2^2$. Vector quantization loss ($\mathcal{L}_{VQ}$) encourages encoder outputs to commit to the codebook entries, defined as $\mathcal{L}_{VQ} = \|sg(z_e) - z_q\|_2^2 + \beta\|z_e - sg(z_q)\|_2^2$, where $sg(.)$ denotes a stop-gradient operator. Adversarial loss ($\mathcal{L}_{adv}$) ensures perceptual motion realism through a discriminator $D$, formulated as $\mathcal{L}_{adv} = -E_{\hat{X}}[D(\hat{X})]$. Together, these losses train the encoder, codebook, and decoder jointly to produce a discrete latent representation that preserves motion dynamics, forming a compact vocabulary of normal motion that serves as the foundation for anomaly detection. During inference, a new keypoint sequence is passed through the encoder-quantizer-decoder pipeline to obtain its reconstruction. Since VQ-VAD is trained exclusively on normal motion, the learned codebook effectively covers only typical behavior patterns. Normal motions can be reconstructed accurately, resulting in low errors. In contrast, abnormal behaviors cannot be represented well using the available motion tokens and therefore produce large reconstruction errors. Consequently, the reconstruction error acts as the anomaly score. For an input sequence $X$ and its reconstruction $\hat{X}$, the anomaly score is defined as $\|X - \hat{X}\|_2$. A final anomaly decision is obtained by thresholding this score for each frame:

Final anomaly decision$(X) = \begin{cases} normal, \ if\ Score(X) < \tau \\ anomaly, \ if\ Score(X) \geq \tau \end{cases}$. In multi-person scenes, each individual is processed independently, and the frame-level anomaly score is computed as $max_i \|X_{i,t} - \hat{X}_{i,t}\|_2$. Thus, if any single person deviates from the normal motion vocabulary, the entire frame is flagged as anomalous. Regarding threshold selection, several strategies commonly used in the literature [7, 20], such as Equal Error Rate (EER), best $F_1$, and $H_{prs}$ threshold, can be employed for real-world deployment.

# 4. EXPERIMENTS

We design our experiments to answer two key questions:

(1) Can VQ-VAD learn a compact and expressive representation of human motion from pose sequences?

(2) Can the learned discrete motion tokens be effectively used for anomaly detection?

Accordingly, experiments are divided into two stages:

- Representation Learning Stage: We first validate whether VQ-VAD can be used to learn a meaningful and compact representation of human motion from 2D/3D keypoint sequences.
- Anomaly Detection Stage: Evaluating VQ-VAD under cross-domain, in-domain, and cross-dataset settings across multiple benchmarks.

## 4.1. Representation Learning on Human Motion

Before anomaly detection, we evaluate VQ-VAD's ability to learn compact motion representations from 2D and 3D keypoint sequences by analyzing reconstruction quality under different compression levels and codebook sizes.

### 4.1.1. Dataset for Representation Learning

We train VQ-VAD on the CMU Panoptic Dataset [14], which provides diverse synchronized 2D and 3D human motion data from multi-view recordings. We selected 10 cameras and segment videos into overlapping 48-frame windows, yielding 14,081 training (1,267,376 pose frames) and 2,672 validation motion segments.

### 4.1.2. Compression Strategy and Codebook Capacity

Compression is applied only along the temporal dimension using factors F4, F8, and F16 to study the trade-off between compression and motion preservation. To isolate compression effects, the codebook size K is fixed during this analysis. We then vary the codebook size while keeping compression constant to evaluate how representational capacity affects motion discretization and reconstruction quality.

### 4.1.3. Human Representation Learning Results

VQ-VAD was trained in PyTorch on three NVIDIA RTX A6000 using AdamW (a learning rate of $2.25 \times 10^{-5}$, $\beta_1 = 0.5$, and $\beta_2 = 0.9$). The input consists of 2D pose sequences with 18 joints (36 channels). Training uses reconstruction, perceptual, and adversarial losses with weights 1.0, 1.0, and 0.5. Evaluation on the CMU Panoptic Dataset [14] measures reconstruction quality and motion representation using three metrics on 2D keypoint sequences.

- Mean Per-Joint Position Error (MPJPE) measures spatial accuracy by computing the average Euclidean distance between the ground-truth and reconstructed joint positions across all frames and joints = $MPJPE = \frac{1}{FK} \sum_{t=1}^{F} \sum_{j=1}^{K} \left\| X_{t,j} - \hat{X}_{t,j} \right\|_2$ ,
- L1 reconstruction error provides a coordinate-wise measure of reconstruction accuracy by averaging the absolute difference between the ground-truth and reconstructed x- and y-locations, $L1 = \frac{1}{2fK} \sum_{t=1}^{F} \sum_{j=1}^{K} \sum_{c=1}^{2} \left| X_{t,j,c} - \hat{X}_{t,j,c} \right|$,
- Temporal smoothness (TS) quantifies how well the model preserves motion continuity. For each joint, we compute frame-to-frame 2D displacements and compare their average magnitudes for the ground truth and reconstruction. A smaller smoothness gap indicates that the reconstructed sequence exhibits realistic motion dynamics comparable to the ground-truth trajectories,

$$TS = \left| \frac{1}{(F-1)\,K} \sum_{t=1}^{F-1} \sum_{j=1}^{K} \left( \left\| \Delta X_{t,j} \right\|_2 - \left\| \Delta \hat{X}_{t,j} \right\|_2 \right) \right|, \Delta X_{t,j} = X_{t+1,j} - X_{t,j}.$$

In addition to these reconstruction metrics, we also report the VQ-VAD quantization loss, which measures how well the encoder outputs align with the learned codebook and reflects the quality and stability of the discrete motion tokens. Table 1 shows that, as expected, lower temporal compression and larger codebooks improve reconstruction quality. The F4 models achieve the best results, with F4-K1024 obtaining the lowest MPJPE (0.072227) and L1 error (0.045956), outperforming F8 and F16 settings. Increasing the codebook size from 256 to 1024 consistently improves reconstruction, highlighting the importance of rich motion token representations for capturing diverse human motion patterns. We limit the codebook size to 1024, as larger codebooks greatly increase computational cost while providing only marginal reconstruction gains. Since F4-K1024 achieves the best overall performance, we use this configuration for all anomaly detection experiments.

Table 1. Reconstruction performance of VQ-VAD models across different temporal compression rates and codebook sizes. Lower values indicate better reconstruction quality.

| Model | Compression | Codebook Size | MPJPE | L1 | TS | Q-loss |
|---|---|---|---|---|---|---|
| $F4 - K2556$ | 4 × | 256 | 0.104681 | 0.066258 | 0.005183 | 0.055883 |
| $F4 - K512$ | 4 × | 512 | 0.118607 | 0.074889 | 0.004948 | 0.0951 |
| $F8 - K512$ | 8 × | 512 | 0.112625 | 0.071657 | 0.008443 | 0.085784 |
| $F16 - K512$ | 16 × | 512 | 0.125169 | 0.079279 | 0.007158 | 0.045579 |
| $F4 - K1024$ | 4 × | 1024 | 0.072227 | 0.045956 | 0.004999 | 0.00929 |
| $F8 - K1024$ | 8 × | 1024 | 0.105588 | 0.067035 | 0.010357 | 0.105121 |
| $F16 - K1024$ | 16 × | 1024 | 0.115638 | 0.073544 | 0.007519 | 0.073238 |

## 4.2. Anomaly Detection Using VQ-VAD

After validating VQ-VAD's motion representation capability, we evaluate anomaly detection under cross-domain, in-domain, and cross-dataset settings. Experiments are conducted on SHT [15], HR-SHT [15], HuVAD [7], and NWPUC [3], covering diverse surveillance environments and abnormal human activities. Table 2 summarizes the statistics of the anomaly detection datasets used in the experiments.

To assess how well the model distinguishes between normal and anomalous behaviors, we use AUC-ROC, the standard evaluation metric for video anomaly detection tasks. AUC-ROC measures how well the model separates normal from abnormal motion by comparing the True Positive Rate (TPR) and False Positive Rate (FPR) across thresholds. Higher AUC-ROC values indicate stronger anomaly discrimination capability.

Table 2. Statistics of the anomaly detection datasets used in our experiments, showing the number of pose frames in the training and test sets, along with the number of available camera views.

| Datasets | Train | Test | Cameras |
|---|---|---|---|
| SHT [15] | 257,650 | 37,845 | 13 |
| NWPUC [3] | 715,901 | 284,228 | 43 |
| HuVAD [7] | 4,467,271 | 729,404 | 7 |

### 4.2.1. Cross-domain Generalization

To evaluate cross-domain generalization, VQ-VAD is trained only on normal motion sequences from the CMU Panoptic dataset [14] and directly tested on SHT [15], HR-SHT [15], HuVAD [7], and NWPUC [3] without retraining or adaptation. As shown in Table 3, performance varies across datasets, with higher accuracy (70–85%) on SHT [15] and lower results (55–70% ) on the more challenging HuVAD [7]. VQ-VAD demonstrates strong cross-domain generalization despite no exposure to target domains during training. It achieves 75.09% AUC-ROC on SHT [15], 76.69% on HR-SHT [15], 61.54% on NWPUC, and 57.4% on HuVAD [7], remaining competitive with several in-domain methods. These results show that VQ-VAD learns transferable motion primitives from the CMU Panoptic dataset [14], enabling robust anomaly detection under domain shift without retraining.

Table 3. Comparison of VQ-VAD with state-of-the-art pose-based anomaly detection methods under cross-domain and in-domain settings.

| Methods | Source | SHT [15] | HR-SHT [15] | NWPUC [3] | HuVAD [7] |
|---|---|---|---|---|---|
| Zaheer et al. [23] | CVPR | 78.93 | - | - | - |
| Hasan et al. [10] | CVPR | 70.40 | 69.80 | - | - |
| Liu et al. [15] | CVPR | 72.80 | 72.70 | - | - |
| MPED-RNN [18] | CVPR | 73.40 | 75.40 | - | 76.05 |
| GEPC [17] | CVPR | 75.50 | - | 62.04 | 62.25 |
| PoseCVAE [13] | ICPR | 74.90 | 75.70 | - | - |
| MSTA-GCN [4] | JVCIR | 75.90 | - | - | - |
| MTP [21] | WACV | 76.30 | 77.04 | - | - |
| STGformer [12] | CVPR | 82.90 | 86.97 | - | - |
| TSGAD [19] | WACV | 80.67 | 81.77 | - | 68.00 |
| STG-NF [11] | ICCV | 85.90 | 87.40 | 62.56 | 57.57 |
| MoPRL [16] | TCSVRT | 83.35 | 84.40 | 61.92 | - |
| **VQ-VAD (Cross-domain)** | - | 75.09 | 76.69 | 61.54 | 57.40 |
| **VQ-VAD (In-domain)** | - | 80.55 | 81.83 | 61.72 | 65.33 |

### 4.2.2. In-Domain Evaluation

To further evaluate VQ-VAD under in-domain setting, we train and test VQ-VAD on the same anomaly detection dataset. Unlike the previous cross-domain generalization experiment, this setup allows us to assess how effectively the model can learn dataset-specific motion patterns when both training and testing occur within the same domain. As shown in Table 3, performance varies across datasets depending on the dataset characteristics. VQ-VAD achieves an AUC-ROC of 80.55% on SHT [15] and 81.83% on HR-SHT [15], demonstrating competitive or superior performance compared to recent pose-based anomaly detection methods. While STG-NF [11] achieves the highest AUC-ROC on these two datasets, a notable advantage of VQ-VAD appears on the HuVAD [7] dataset, one of the most challenging benchmarks, it reaches 65.33%, significantly outperforming STG-NF [11] (57.57%). This result indicates that VQ-VAD's motion tokenization handles the increased difficulty and variability of complex real-world scenes well. Additionally, on the NWPUC [3] dataset, all pose-based methods with available results achieve accuracy in the 61-63% range, and VQ-VAD's AUC-ROC aligns closely with this trend.

### 4.2.3. Cross-dataset Generalization

To evaluate cross-dataset generalization, VQ-VAD and STG-NF [11] are trained on SHT [15] and directly tested on NWPUC [3] and HuVAD [7] without adaptation. On NWPUC [3], both models achieve similar performance (61.81% for VQ-VAD vs. 62.44% for STG-NF [11]). However, on the more challenging HuVAD [7], VQ-VAD achieves 59.66%, outperforming STG-NF [11] at 54.32% by more than 5%. These results highlight the stronger robustness of VQ-VAD's discrete motion representations under domain shift.

Table 4. Cross-dataset generalization results of VQ-VAD and STG-NF [11].

| **Models** | **Tested on NWPUC [3]** | **Tested on HuVAD [7]** |
|---|---|---|
| STG-NF [11] | 62.44 | 54.32 |
| VQ-VAD | 61.81 | 59.66 |

VQ-VAD prioritizes stability and cross-domain robustness over peak in-domain accuracy. While methods such as STG-NF [11] achieve higher AUC on SHT [15], they rely more on scene-specific features and generalize poorly across datasets. In contrast, VQ-VAD learns a discrete motion vocabulary that captures transferable motion semantics. Across in-domain and cross-dataset evaluations, VQ-VAD achieves competitive performance on SHT [15] and HR-SHT [15], while consistently outperforming STG-NF [11] on the more challenging HuVAD [7] under domain shift. These results highlight the robustness and generalization capability of VQ-VAD for real-world anomaly detection.

# 5. CONCLUSION

In this work, we introduced VQ-VAD, a vector-quantized framework for human-centric video anomaly detection that models motion through discrete latent representations. By adapting VQ-GAN to temporal pose sequences, VQ-VAD learns a compact and expressive motion vocabulary for anomaly detection. Experiments on the CMU Panoptic dataset [14] show strong motion reconstruction under temporal compression, while evaluations across cross-domain, in-domain, and cross-dataset settings demonstrate robust and transferable anomaly detection performance. These results highlight VQ-VAD as an effective and generalizable framework for human-centric VAD.

# ACKNOWLEDGEMENTS

This research is supported by the National Science Foundation (NSF) under Award Numbers 2329816.